\documentclass[conference]{IEEEtran}
\usepackage{amsmath,graphicx}
\usepackage{algorithm} 
\usepackage{url} 
\usepackage{algorithmic}  
\usepackage{booktabs}
\usepackage{subfigure}
\usepackage{enumitem}
\setlist{nosep, leftmargin=14pt}
\usepackage{mwe} 

\begin{document}

\title{MedISure: Towards Assuring Machine Learning-based Medical Image Classifiers using Mixup Boundary Analysis}

\author{Adam Byfield$^1$, William Poulett$^1$, Ben Wallace$^1$, Anusha Jose$^1$, Shatakshi Tyagi$^1$, Smita Shembekar$^1$, \\ {Adnan Qayyum}$^2$, {Junaid Qadir}$^3$, {and Muhammad Bilal}$^{4,*}$\thanks{Email: Muhammad.Bilal@bcu.ac.uk} \\
$^1$ NHS England (National Health Service), UK \\
$^2$ Information Technology University of the Punjab, Lahore, Pakistan\\
$^3$ Qatar University, Doha, Qatar \\
$^4$ Birmingham City University (BCU), Birmingham, UK}

\maketitle            

\begin{abstract}

Machine learning (ML) models are becoming integral in healthcare technologies, presenting a critical need for formal assurance to validate their safety, fairness, robustness, and trustworthiness. These models are inherently prone to errors, potentially posing serious risks to patient health and could even cause irreparable harm. Traditional software assurance techniques rely on fixed code and do not directly apply to ML models since these algorithms are adaptable and learn from curated datasets through a training process. However, adapting established principles, such as boundary testing using synthetic test data can effectively bridge this gap. To this end, we present a novel technique called Mix-Up Boundary Analysis (MUBA) that facilitates evaluating image classifiers in terms of prediction fairness. We evaluated MUBA for two important medical imaging tasks---brain tumour classification and breast cancer classification---and achieved promising results. This research aims to showcase the importance of adapting traditional assurance principles for assessing ML models to enhance the safety and reliability of healthcare technologies. To facilitate future research, we plan to publicly release our code for MUBA. 

\end{abstract}

\section{Introduction}
\label{sec:intro}
Recent advances in Artificial Intelligence (AI) have resulted in a rapid uptake of AI-based algorithms in healthcare with applications in areas including radiology, pathology, and complex surgery. The high-risk nature of the industry and the prospective impact (positive or adverse) of AI algorithms on the lives of patients means they must be rigorously tested before deploying them for use by clinicians \cite{neto2022safety}. These healthcare professionals can only be confident in AI-empowered healthcare tools if they are cognisant of the fact that they have been developed and tested using a proven and robust methodology. Unfortunately, traditional software testing and assurance techniques can not be directly applied to AI models as these systems are data-intensive, self-adapting, and continuously evolving \cite{martinez2022software}. Hasty deployment of AI-based solutions in healthcare without reliable assurance can result in a heavy price of compromising patient safety \cite{batarseh2021survey}.

The AI community currently creates and shares summary statistics to assess model performance for quality assurance \cite{fujii2020guidelines}. These statistics provide a good starting point to empirically justify the model's performance. However, these alone are insufficient to verify the models' behavioural subtleties. It is likely that in the pursuit of improving the model's accuracy, one might reintroduce previously fixed bugs, or incorporate nuanced issues like gender bias in the model's behaviour. AI assurance involves measuring, evaluating, and communicating various information from model training through validation up to deployment and monitoring while ensuring governance and compliance with laid-out standards and regulations. AI assurance is largely uncharted territory for healthcare solutions and a robust AI assurance methodology for Deep Learning (DL)-based medical solutions is lacking.

\begin{figure}[!t]
    \centering
    \includegraphics[width=0.48\textwidth]{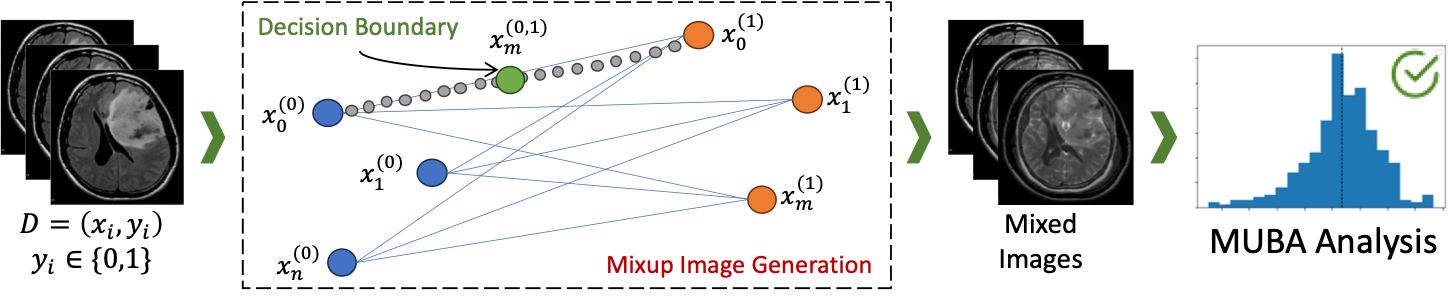}
    \caption{Interpreting decision boundaries of ML-based binary image classifiers using mix-up images (an image $x_i^{(c)}$ is overlayed on $x_j^{(c)}$, where class $c \in \{0,1\}$) and MUBA histograms.} 
    \label{fig:method}
\end{figure}

Many healthcare datasets are plagued by issues like substantial class imbalances. This can lead to the development of models that are both biased and unreliable \cite{qayyum2020secure}. When such models are applied in clinical settings, where clinicians make critical care decisions, the resulting unjust AI predictions could endanger patient lives. Recent research emphasises the essential requirement for robust internal and external validation measures for AI models in medicine \cite{ghassemi2021false}. Algorithmic audit techniques play a crucial role in revealing the limitations of models and enhancing trust in their predictions. Xiaoxuan et al. \cite{liu2022medical} have outlined a comprehensive framework for medical algorithmic audit, detailing processes and documents produced by various stakeholders in the product design. While traditional software assurance provides established principles that can be adapted for ML assurance, there is a scarcity of studies applying these principles. This study addresses this gap by demonstrating the application of boundary testing to gain a deeper understanding of model behaviour. As illustrated in Fig. \ref{fig:method}, we put forward Mix-Up Boundary Analysis (MUBA)---a novel interpretability method for identifying the model's decision boundary for assuring medical image classification models.
Our method utilizes mixed-up images (a combination of class 0 and 1 images), to establish the model's decision boundary. For example, the boundary point for a pair of original images is identified by observing the change in the model's prediction when using their mixed-up counterparts (depicted as a green point in Fig. \ref{fig:method}). Specifically, the following are the salient contributions of this paper.

\begin{enumerate}
    \item We present MUBA, a novel interpretability method to evaluate the prediction boundaries of binary classification models. 
    \item We empirically evaluate the use of MUBA analysis in assuring DL-based binary and multi-class classifiers.   
\end{enumerate}

\section{Mix-Up Boundary Analysis (MUBA)}
\label{sec:method}

\subsection{Mixup Image Generation}
\begin{figure}[!t]
    \centering
    \includegraphics[width=0.49\textwidth]{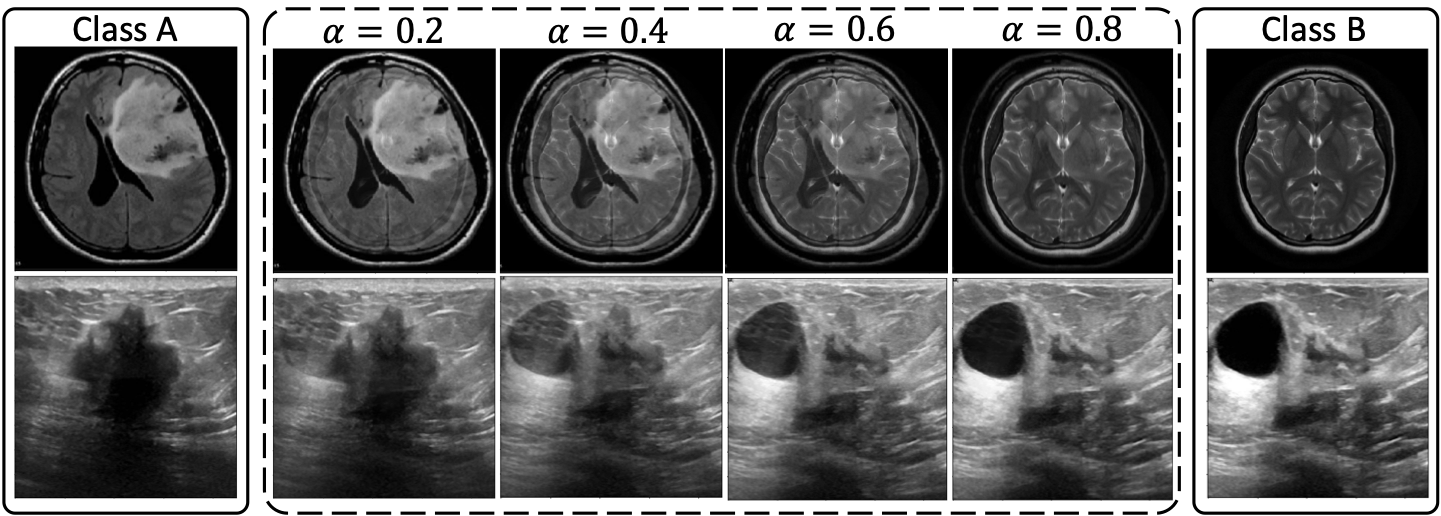}
    \caption{Mixed-up images for different $\alpha$ values (dotted box) using original images of two classes (solid box).}
    \label{fig:mixed}
\end{figure}

In traditional software testing, boundaries often harbor defects and thus are scrutinised rigorously to confirm that the software behaves correctly in these areas. We apply this analogy to the case of testing a binary image classifier where there is an effective decision boundary between the two classes, i.e., a tipping point at which the model will move from predicting one class to the other. To achieve this, we proposed a novel technique MUBA, which is inspired by the well-known mix-up augmentation technique to train image classifiers \cite{zhang2017mixup}. For a binary classifier with class sample sizes of $n$ and $m$, $N$ mixed-up images are generated between each data point in both classes, resulting in $N \times n \times m$ total mixed-up images. The generation of mixed-up images from two classes, i.e., ($x_i^{(0)},y_i^{(0)}$) and ($x_j^{(1)},y_j^{(1)}$) from the training data can be mathematically expressed as: 

\begin{equation}
    x_{m}^{(0,1)} = \alpha \times (x_i^{(0)},y_i^{(0)}) + (1 - \alpha) \times (x_j^{(1)},y_j^{(1)}),
\end{equation}

where $\alpha$ is the random mixing coefficient, selected from a uniform distribution within the bin size of $1/N$, ensuring an even distribution of $\alpha$ values per mix-up image $x_{m}^{(0,1)}$. Mixed images are then used to infer the models to get insight into the model's decision boundary. Examples of mixed-up images using two classes can be seen in Fig. \ref{fig:mixed}.  

\subsection{MUBA Histograms Generation and Interpretation}
Deep neural networks are vulnerable to small perturbations of the input data, which can completely change the predicted label \cite{qayyum2020secure}. These can be intentionally created adversarial examples but can also be caused by distributional shifts or image noise \cite{volpi2019addressing}. Mixed-up images are a direct mix of features from two separate classes, therefore, perturbations pull the image towards the decision boundary, even in latent space. A decision boundary at $\alpha = 0.5$ has the lowest risk of misclassifying test images with small perturbations and hence is the most robust. When predictions change between two mix-up images, the middle $\alpha$ value between the two images is considered as a boundary point. These values can be plotted as a histogram to generate a Boundary Distribution Curve (BDC). For this reason, we expect BDCs to be centred at $\alpha = 0.5$. Moreover, incorrectly classified images have been shown to lie on average closer to the decision boundary \cite{mickisch2020understanding}. Similarly, we count the number of incorrect predictions across different $\alpha$ values to produce an Error Distribution Curve (EDC). These curves both have the benefit of working on black box models where only the final predicted label is known. If a model predicts both classes equally well, we expect both curves to be approximately symmetrical. 

\section{Results and Discussions}
\label{sec:results}

\subsection{Data Description and Experimental Setup}
To evaluate MUBA, we used Brain Tumour\footnote{\url{https://tinyurl.com/582sfhmb}} and Breast Cancer classification\footnote{\url{https://tinyurl.com/mt5pkm9f}} datasets that contain labelled images for binary and three-class classification problems. Specifically, the brain tumour dataset consists of 253 images (i.e., 98 images in normal and 155 in tumour class), and the breast cancer dataset contains 891, 421, and 266 in benign, malignant, and normal classes, respectively. To address the class imbalance issue, we performed data augmentation of minority classes using random horizontal flips and rotations. We used ResNet152V2 pretrained on ImageNet as our baseline and stacked two convolutional and two fully connected on top of the base model (before the classification layer). Furthermore, to overcome the overfitting issue and to ensure stable training, we added batch normalisation layers after these newly added layers. These layers were only finetuned using our custom datasets using a split of 70\%, 15\%, and 15\%, for training, validation, and testing. We first finetuned the model using a learning rate of 0.001 and batch size of 32 for a maximum of 60 epochs, which was further finetuned for 60 more epochs using a learning rate of 0.0001 (where the last 100 layers were tunable, unlike the initial version of the model (where only new layers were finetuned). We keep the best-performing model for analysis using the MUBA method. We used TensorFlow ML library and Python for our implementation.         

\begin{figure}[!t]
    \centering
    \includegraphics[width=0.48\textwidth]{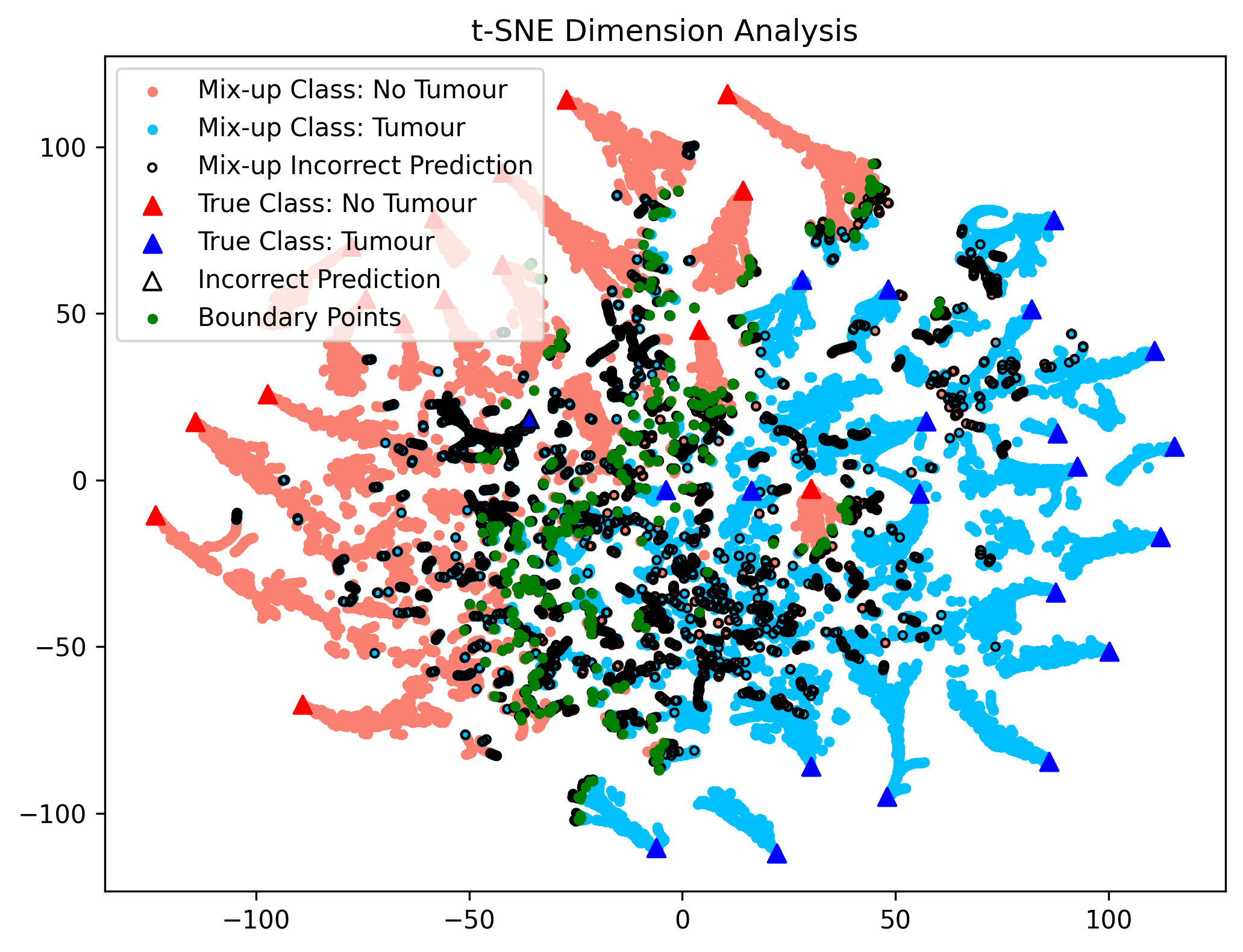}
    \caption{Demonstrating effectiveness of mixed-up images to interpret decision boundary (for brain tumour classification).}
    \label{fig:tSNE}
\end{figure}

\subsection{Interpreting Decision Boundaries using MUBA}
We first trained a best-performing baseline model for the binary brain tumour classification, achieving an accuracy of 97\% and an F1 score of 0.98. Secondly, we generated mixed-up images using our test set of 38 images to infer the trained model and demonstrate intuitive reasoning for MUBA. Precisely, 60 mix-up images were generated using random $\alpha$ values between each of the 17 images in the ``no tumour'', and the 21 in the ``tumour'' class within a bin of width $1/60$ to ensure an even distribution. Thirdly, we extracted the embeddings of the underlying model from the last layer (before the classification layer) and used t-SNE to visualise high-dimensional embeddings corresponding to inference images in 2D space. Fig. \ref{fig:tSNE} illustrates the embedding space for model predictions using original and mixed images including correct and incorrect predictions. We can see from the figure that predictions using mixed-up images filled the gap between the true classes in the latent embedding space. The green points in Fig. \ref{fig:tSNE}, indicate the model's decision boundary point (these points refer to the the mixed-up images where the predicted label was changed). This indicates that although mixed-up images are not necessarily the same as true test images, they can help identify or add detail to the actual decision boundaries within latent space. For instance, we see incorrect predictions generally around the model's decision boundary.

\subsection{Model Evaluation using MUBA Histograms}
The dimensionality reduction analysis showed that the distribution of errors and the mixed-up images where the model changes its decision can be used to describe the model's behaviour around the boundary in the latent space. Therefore, plotting these predictions in histograms can provide the distribution of errors and boundary points with respect to the given value of $\alpha$. Below we present the analysis of binary (brain tumour) and multi-class (breast cancer) image classifiers using MUBA histograms that include BDC and EDC. 

\begin{figure}[!t]
    \centering
    \subfigure[]{\includegraphics[width=0.48\textwidth]{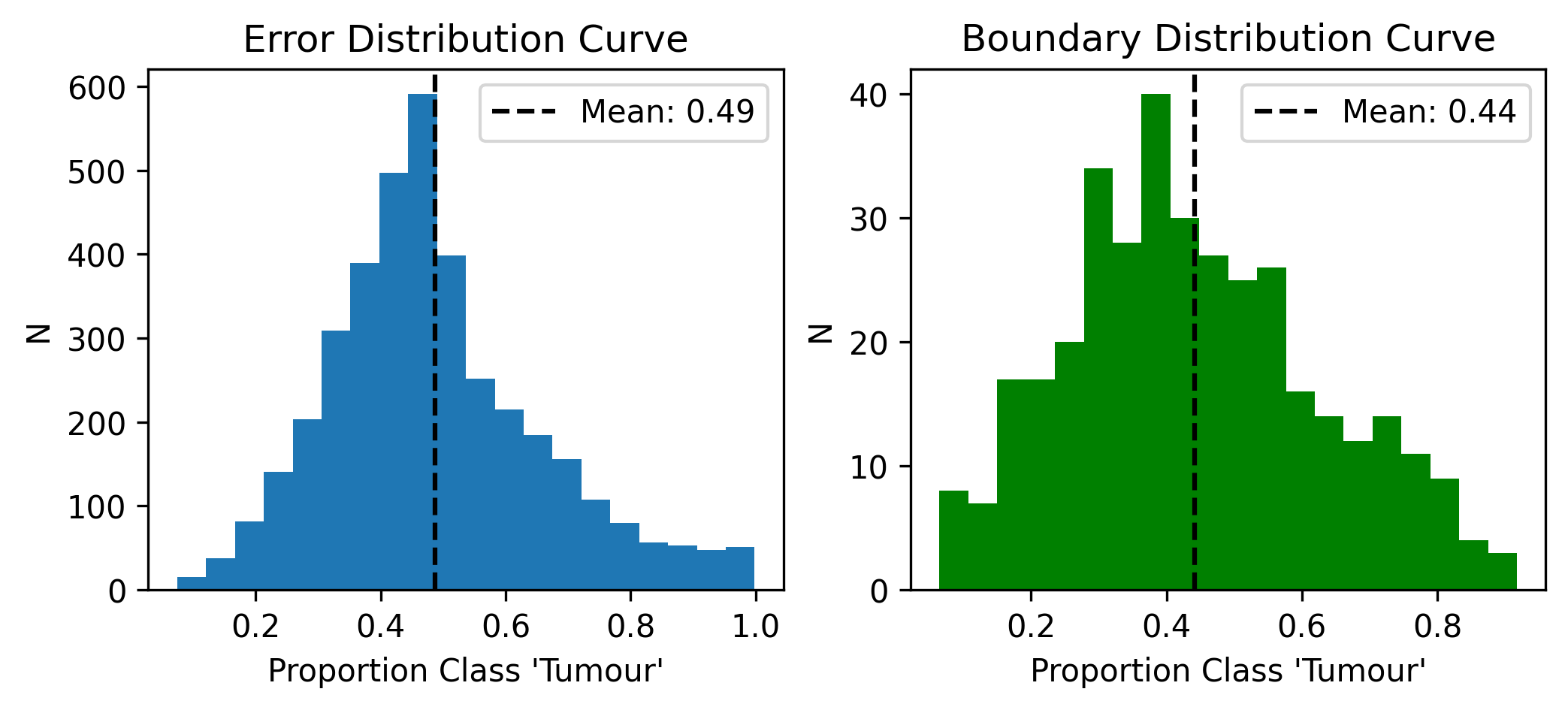}\label{fig:bds_good}}
    \subfigure[]{\includegraphics[width=0.48\textwidth]{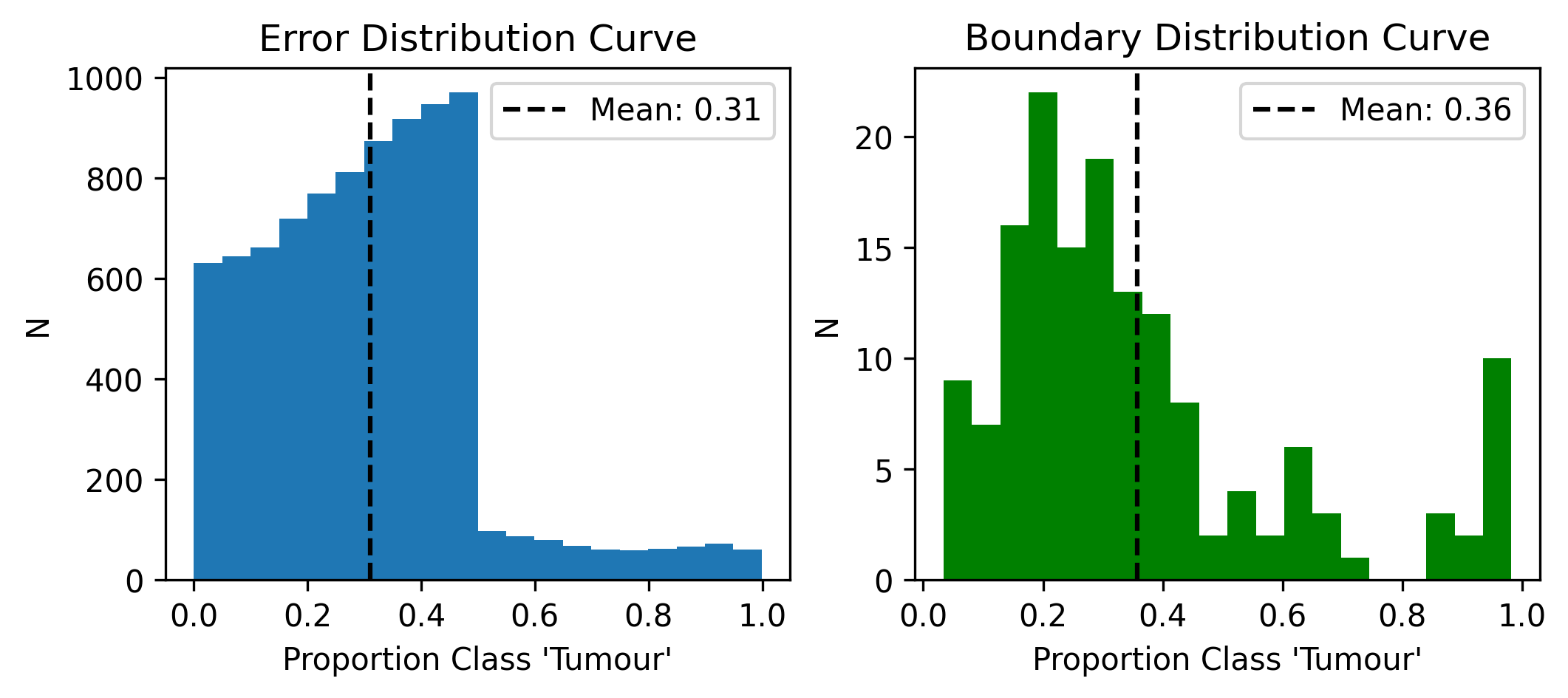}\label{fig:bds_bad}}
    \caption{Boundary distribution curve (BDC) and error distribution curve (EDC) for a 97\% (Fig. \ref{fig:bds_good}) and 71\% accurate binary brain tumour classification model (Fig. \ref{fig:bds_bad}).}
    \label{fig:bds_good_vs_bad}
\end{figure}

\begin{figure*}[!t]
    \centering
    \includegraphics[width=\textwidth]{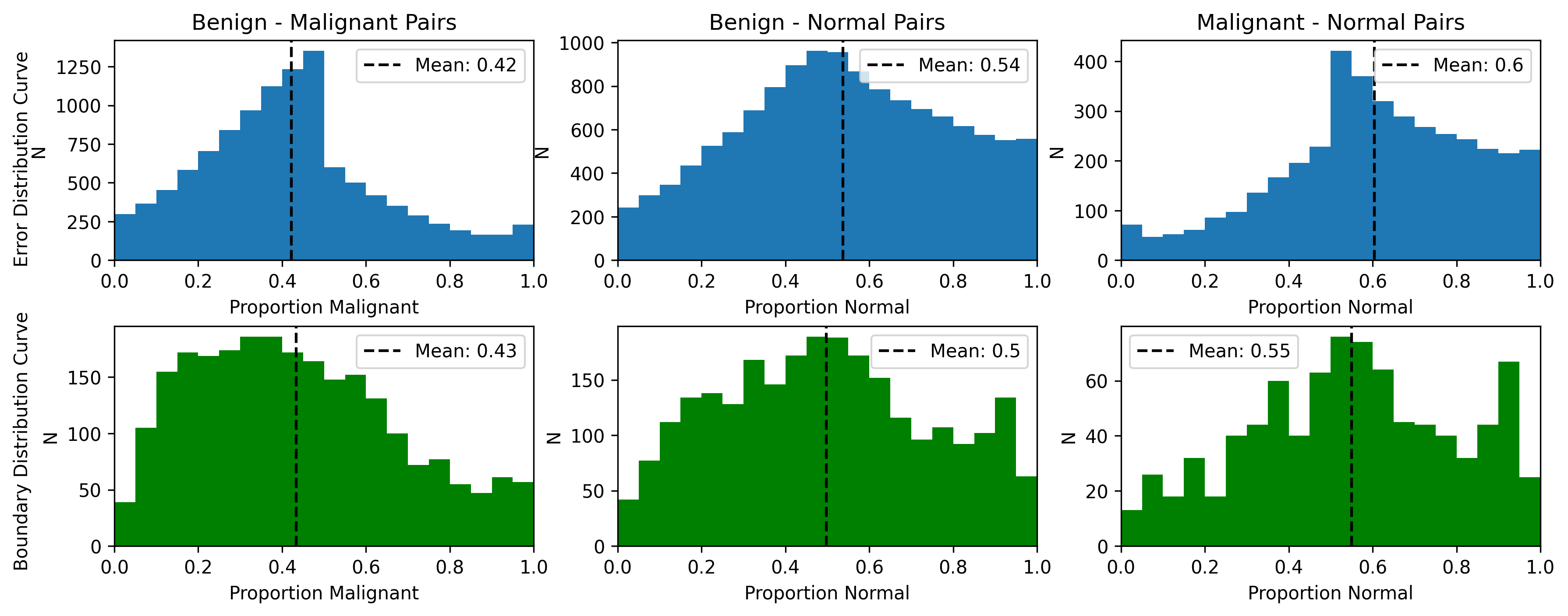}
    \caption{EDC curves (top) and BDC curves (bottom) for an 80\% accurate three-class breast cancer classifier.}
    \label{fig:3_class_muba}
\end{figure*}

\textit{Brain Tumour Classification:} Fig. \ref{fig:bds_good_vs_bad} shows MUBA histograms (i.e., BDC and EDC) for two sets of brain tumour classification models having an accuracy of 97\% (Fig. \ref{fig:bds_good}) and 71\% (Fig. \ref{fig:bds_bad}). It is evident from Fig. \ref{fig:bds_good} that errors are distributed normally, with the most errors occurring around the $\alpha$ value of $0.5$. Also, we can see that there are comparatively more errors on the left side of the histograms; this is where the images were mostly ``without tumour" but were being classified as ``tumour". Moreover, Fig. \ref{fig:bds_good_vs_bad} shows that the model mainly changes its predictions slightly to the left of the $\alpha = 0.5$ mark, which indicates a propensity to say an image contains a tumour when it does not. On the other hand, for a 71\% accurate model (Fig. \ref{fig:bds_bad}, we see a mean boundary closer to $\alpha = 0.3$, indicating a bias towards predicting an image as to having no tumour; therefore, the boundary appears to be closer to the ``no tumour" class. Moreover, it is evident from Fig. \ref{fig:bds_good_vs_bad} that the EDC of a model performing reasonably well is more symmetrical than the model with less performance (e.g., see Fig. \ref{fig:bds_bad}). 

\textit{Breast Cancer Multi-class Classification:} In addition to evaluating MUBA for binary image classifiers, we also assess its efficacy against three-class classification models (i.e., breast cancer classification). For instance, in Fig. \ref{fig:3_class_muba}, we demonstrate that EDC and BDC can be effectively generated by considering a pair of classes from $N$ total classes. The figure shows the MUBA histograms for an 80\% accurate model trained using a breast tumour dataset containing three classes. Fig. \ref{fig:3_class_muba} reveals that the mean boundary for errors in EDC plots is shifted towards the ``normal" class due to the high proportion of ``normal" class images. Also, we see that for pairs containing the ``normal" class, BDC shows a mean boundary closer to $\alpha = 0.5$. Moreover, there are spikes in the BDC curves between $\alpha = 0.8$ and $\alpha = 1$ for images with a high proportion of ``normal". This indicates that some images incorrectly labelled as ``normal" do not need a high proportion of another class to be predicted differently. This insight into a model's decision boundary without knowing the confidence of a model's prediction is unique to MUBA.

\section{Conclusions}
\label{sec:cons}
This paper presents a Mix-up Boundary Analysis (MUBA) that leverages mix-up augmentation to fortify machine learning (ML) models during the training phase, particularly by challenging the decision boundaries with synthetic test data. We show that images synthesised from mixed classes inhabit the models' latent decision space, providing a new means for boundary testing. MUBA does not seek to replace but rather complement quality testing by exposing the model to intricate synthetic test images that may reveal biases, especially around decision boundaries. Notably, MUBA's methodology is uniquely beneficial because it requires no access to the internal model weights or confidence scores, thus offering a distinct advantage for third-party assurance in sensitive fields such as medical image classification. Our study confirms that while traditional software testing strategies are insufficient for assuring ML-based medical image classification systems, the systematic application of MUBA can be used for algorithmic audit, thereby enhancing the understanding, performance, and fairness of predictions in these models. Moreover, our research highlights the critical need for collaboration between traditional software assurance professionals and ML experts to innovate testing methods, paving the way for the broader adoption and trust in ML within healthcare specialities.

\bibliographystyle{IEEEtran}

\end{document}